\documentclass[10pt,twocolumn,letterpaper]{article}

\usepackage{iccv}
\usepackage{times}
\usepackage{epsfig}
\usepackage{graphicx}
\usepackage{amsmath}
\usepackage{amssymb}
\usepackage{multirow}
\usepackage{booktabs} 
\usepackage{diagbox}
\usepackage{algorithm}
\usepackage{algorithmic}
\usepackage[table]{xcolor}

\usepackage[pagebackref=true,breaklinks=true,letterpaper=true,colorlinks,bookmarks=false]{hyperref}

\iccvfinalcopy 

\def\iccvPaperID{****} 

\ificcvfinal\fi

\begin{document}

\title{On the Effect of Pruning on Adversarial Robustness}

\author{Artur Jordao and Helio Pedrini\\
Institute of Computing, University of Campinas (UNICAMP), Brazil\\
{\tt\small  arturjlcorreia@liv.ic.unicamp.br, helio@ic.unicamp.br}
}

\maketitle
\ificcvfinal\thispagestyle{empty}\fi

\begin{abstract}
Pruning is a well-known mechanism for reducing the computational cost of deep convolutional networks.
However, studies have shown the potential of pruning as a form of regularization, which reduces overfitting and improves generalization. 
%
We demonstrate that this family of strategies provides additional benefits beyond computational performance and generalization.
Our analyses reveal that pruning structures (filters and/or layers) from convolutional networks increase not only generalization but also robustness to adversarial images (natural images with content modified).
Such achievements are possible since pruning reduces network capacity and provides regularization, which have been proven effective tools against adversarial images.
%
In contrast to promising defense mechanisms that require training with adversarial images and careful regularization, we show that pruning obtains competitive results considering only natural images (e.g., the standard and low-cost training).
%
We confirm these findings on several adversarial attacks and architectures; thus suggesting the potential of pruning as a novel defense mechanism against adversarial images.
\end{abstract}
\section{Introduction}\label{sec:introduction}
\begin{figure}[!t]
\centering
\includegraphics[width=\columnwidth]{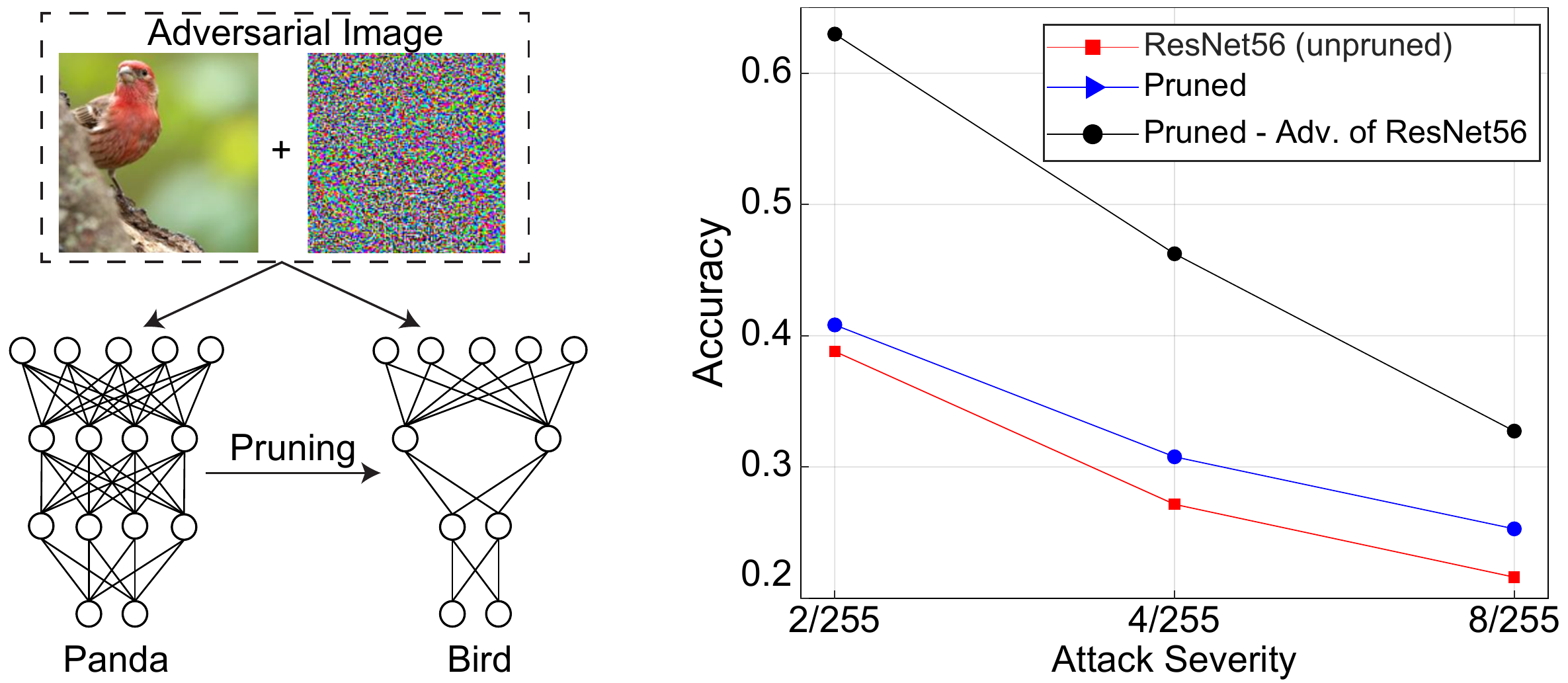}
\caption{Predictive behavior of an over-parameterized network and its pruned counterpart on adversarial images. In this example, adversarial images are crafted by Fast Gradient Sign Method (FGSM). \textbf{Left}. Forwarding an adversarial image through the high-capacity (i.e., over-parameterized) network results in a wrong prediction  -- the network classifies the bird image as a panda. After removing its neurons, the network classifies the image correctly, suggesting that the pruned network became less vulnerable to the attack. \textbf{Right.} Accuracy on adversarial images (robustness) of ResNet56 and its pruned versions. On the standard FGSM attack, in which a network is attacked by adversarial images crafted from its parameters, the pruned ResNet56 (blue curve) achieves better robustness than its unpruned (red curve) counterpart. The robustness is even better when the adversarial images are yielded by ResNet56 and used to attack its pruned version, i.e., the attack is unaware of pruning (black curve). Note that such findings are confirmed under different attack severities.}
\label{fig:teaser}
\end{figure}

Despite achieving remarkable results in image classification~\cite{Tan:2019:ICML, Kolesnikov:2020}, modern and high-performance convolutional networks are easily fooled by adversarial images~\cite{Hendrycks:2019:ICLR, Hendrycks:2021:CVPR}.
Unlike natural (clean) images, adversarial images have their content modified to force a network to wrong its prediction.
%
Figure~\ref{fig:teaser} (left) shows an adversarial image and its effect on an over-parameterized network (leftmost). Since adversarial images exist in real-world settings, increasing the adversarial robustness of convolutional networks plays a role in safety- and security-critical applications.
%

Many mechanisms provide defense against adversarial images, for example, transfer learning~\cite{Hendrycks:2019:ICML, Shafahi:2020}, regularization~\cite{Xie:2020, Xie:2020:CVPR} and data augmentation~\cite{Chun:2019, Hendrycks:2020:ICLR, Li:2021}. 
Regarding the latter, there is a successful category that combines clean and adversarial images during the training phase, named adversarial training~\cite{Madry:2018}. Despite providing remarkable results, adversarial training requires careful regularization; otherwise, it hurts predictive ability on clean images~\cite{Chun:2019, Raghunathan:2019, Raghunathan:2020, Xie:2020:CVPR, Xie:2020}. 
Moreover, adversarial training might lack effectiveness when training data is scarce~\cite{Raghunathan:2019, Carmon:2019} or the settings of the attack are changed~\cite{Wang:2020}.
%
%

Regardless of the defense mechanisms, most efforts attempt to circumvent the dilemma between adversarial robustness (accuracy on adversarial images) and generalization (accuracy on clean images), which means improving network accuracy on both domains.

According to previous works~\cite{Tran:2018, Kaya:2019, Kuei:2020, Zhou:2020}, the success of adversarial images is due to the excessive over-parameterization and capacity of deep networks. For example, Kaya et al.~\cite{Kaya:2019} and Hu et al.~\cite{Kuei:2020} observed that stopping inference in early layers improves adversarial robustness. Analogously, this mechanism aims at finding a shallow (low capacity) network inside a deeper network (high capacity). Another evidence is that networks with strong regularization are less vulnerable to adversarial images~\cite{Chun:2019}. Building upon these findings, a natural idea is to increase adversarial robustness by reducing network capacity and forcing regularization while preserving generalization. In this work, we show that it is possible to satisfy such conditions by eliminating structures (filters, layers or both) from convolutional networks.

Our analysis relies on the evidence of previous research on compression and acceleration of deep networks, which have demonstrated the potential of pruning as a form of regularization capable of reducing overfitting and improving generalization~\cite{Huang:2016, Li:2019:CVPR, Fan:2020, Bartoldson:2020}. We leverage such behavior through the lens of adversarial images and demonstrate that removing filters and/or layers increases adversarial robustness without hurting generalization. Therefore, this process promotes an effective defense mechanism.

\noindent
\textbf{Contributions.} Our key contribution is to demonstrate that eliminating structures of convolutional networks -- pruning -- increases their adversarial robustness, see Figure~\ref{fig:teaser}~(right).
%
In order to confirm this, we prune different structures composing a convolutional network such as filters, layers and both. Importantly, throughout the pruning process, we take into account only clean images. In practice, this means that we can ignore the settings or any additional assumption of the attack when designing a defense mechanism. Further, it implies that we can avoid careful regularization and use the \emph{low-cost} (standard -- clean images only) training as a defense mechanism.
By pruning layers, we show that shallower networks might obtain superior adversarial robustness than their deeper counterpart. This evidence exhibits an opposite phenomenon to a recent finding, which states that some defense mechanisms require deeper (costly) networks~\cite{Xie:2020, Xie:2020:CVPR}.
%
%
All these findings appear without decreasing accuracy on clean images, which demonstrates the potential of pruning in satisfying the dilemma between generalization and robustness. However, assuming a negligible drop in generalization (to be precise less than one percentage point), we demonstrate that removing a single structure (e.g., only one filter) without any parameter adjustment enables a pruned network to obtain better robustness than its unpruned (defenseless) version.

We validate the claims above on several adversarial attacks (semantic-preserving, Fast Gradient Sign Method and occlusion), convolutional architectures (VGG, MobileNets and Residual networks) and datasets (CIFAR-10 and ImageNet). Additionally, we assess the effectiveness of pruning on many settings such as the structure removed, the pruning criterion and the scheme to adjust the parameters of the networks. On these configurations, pruning obtained gains in adversarial robustness of up to $4.29$ percentage points. These gains substantially increase when we craft adversarial images using a convolutional architecture and attack a pruned version of other architecture.
%
Compared to state-of-the-art defense mechanisms, which often involve training from scratch on adversarial images, pruning obtained competitive results performing few epochs of fine-tuning on clean images. Furthermore, it achieved one of the best trade-offs between robustness and generalization.
%
%
%
%
%
%
%
\section{Preliminaries and Problem Statement}
\noindent
\textbf{Notations.} Define $X$ and $Y$ the set of images and their respective class labels. Let $x \in X$ and $y \in Y$ be a single image and label of $X$ and $Y$, respectively. Let $\mathcal{F}$ be a convolutional network that receives an input image $x$ and outputs $j$ probabilities, each one representing the probability of $x$ belonging to the $jth$ class. 
Thus, the classification of $x$ using $\mathcal{F}$ is $\max(\mathcal{F}(x))$, and $\mathcal{F}$ correctly classifies $x$ if $\max(\mathcal{F}(x))=y$. For simplicity, we will omit the notation $\max(.)$.
%

\noindent
\textbf{Adversarial Attack.} 
For a given image $x$, assume that a network $\mathcal{F}$ correctly classifies $x$, $\mathcal{F}(x)=y$. An adversarial attack crafts an adversarial version of $x$, $x'$, so that $\mathcal{F}$ makes a misclassification, which means $\mathcal{F}(x') \neq y$.
Equation~\ref{eq:corruption} expresses the process of crafting an adversarial image.
\begin{equation}\label{eq:corruption}
x' = x + \alpha\delta, \delta \in \Delta,
\end{equation} 
where $\delta$ represents a single corruption given a set of possible corruptions $\Delta$, which include semantic-preserving transformations and corruptions designed using the own parameters of a network (i.e., model-specific attacks). 
The parameter $\alpha$ controls the severity/intensity of the corruption, for which low values become $x'$ similar to $x$ and visually imperceptible to humans, but the attack is less effective. 
%

\noindent
\textbf{Pruning Structures.}
%
Let $c$ be a criterion (a.k.a pruning criterion) that receives as input a network $\mathcal{F}$ and assigns the importance for each structure (filters or layers) composing $\mathcal{F}$. Define such an operation as $c(\mathcal{F})$.
Let $S$ be a set of scores (i.e., importance) resulting from applying $c(\mathcal{F})$, with each element of $S$ corresponding to the importance of a single structure of $\mathcal{F}$.
Define $I$ a set of indices indicating the $p\%$ lowest score (unimportant) structures of $\mathcal{F}$ based on $S$; thereby, $|I| = \lceil p \times S\rceil$. The pruning algorithm removes the structures of $\mathcal{F}$ indexed by $I$, yielding a pruned network $\mathcal{F'}$. Specifically, $\mathcal{F'}$ is a thinner (removing filters) or shallower (removing layers) version of $\mathcal{F}$. 
After removing structures, the algorithm adjusts the parameters of the resulting network, $\mathcal{F'}$, for some epochs (fine-tuning).
\begin{algorithm}[!t]
	\small
	\caption{Pruning Structures of Convolutional Networks}
	\label{alg::pruning_structures}
	\begin{algorithmic}
		\STATE {\bfseries Input:} Convolutional Network $\mathcal{F}$
		\STATE {\bfseries Input:} Number of Iterations $K$\\
		\STATE {\bfseries Input:} Pruning Criterion $c$\\
		\STATE {\bfseries Input:} Pruning Ratio $p$\\
		\STATE {\bfseries Output:} Pruned Convolutional Network $\mathcal{F'}$
		\FOR{$k=1$ {\bfseries to} $K$}	
		\STATE $S \leftarrow c(\mathcal{F})$ $\triangleright$ Assigns importance for each structure
		\STATE $I \leftarrow p\%$ unimportant structures based on $S$
		\STATE $\mathcal{F'} \leftarrow \mathcal{F} \setminus I $ $\triangleright$ Removes the structures indexed by $I$ 
		\STATE Update $\mathcal{F'}$ via standard fine-tuning (clean images)
		\STATE $\mathcal{F} \leftarrow \mathcal{F'}$
		\ENDFOR
	\end{algorithmic}
\end{algorithm}

One common practice is to remove the structures of $\mathcal{F}$ iteratively. Such a strategy consists of using $\mathcal{F'}$ of the previous iteration as input to the next iteration. This iterative nature implies that the pruned network of the $k$-th iteration has fewer structures than its preceding versions. Algorithm~\ref{alg::pruning_structures} summarizes the pruning process.

\noindent
\textbf{Problem Statement.} Define $Acc_{clean}(.)$ and $Acc_{adv}(.)$ as the metrics that compute the accuracy of a convolutional network on clean (generalization) and adversarial (robustness) images, respectively.
%
In this work, we verify the following statement:
\begin{equation*}
Acc_{adv}(\mathcal{F'}) > Acc_{adv}(\mathcal{F}), Acc_{clean}(\mathcal{F'}) \approx Acc_{clean}(\mathcal{F}).
\end{equation*}

The left part of the statement above allows us to answer the following research question: \emph{Do pruned networks inherit the vulnerability to adversarial images of their unpruned counterpart?} Both sides allow us to answer the following: \emph{Are pruned networks capable of improving robustness while preserving generalization?}
\section{Related Work}
%
\noindent
\textbf{Adversarial Attacks.} 
Adversarial attacks are mechanisms that add visually imperceptible perturbations to natural (clean) images focusing on degenerating the predictive ability of a network. 
These attacks may have access to the network parameters and architecture design~\cite{Goodfellow:2015, Madry:2018}, or only to the network output~\cite{Rezaei:2020}. 
For example, attacks such as Fast Gradient Signed Method (FGSM) and Projected Gradient Descent (PGD) use the gradient of loss to perturb clean images.
%
Without access to network details (i.e., model-agnostic), it is also possible to make a network wrong its predictions.
As demonstrated by Hendrycks et al.~\cite{Hendrycks:2019:ICLR}, semantic-preserving transformations (e.g., blur and contrast) applied to clean images are capable of degenerating the performance of high-capacity networks. According to previous works~\cite{Yun:2019, Chun:2019, Shi:2020}, even simple occlusions in clean images confound the network prediction.
	
\noindent
\textbf{Adversarial Defenses.} Defense mechanisms are techniques that aim at reducing the vulnerabilities of convolutional networks to adversarial attacks. 
An effective defense mechanism is to increase the variability of training samples by data augmentation~\cite{Yun:2019, Chun:2019, Hendrycks:2020:ICLR, Yingwei_Li:2021:ICLR}. 
From an empirical perspective, Chun et al.~\cite{Chun:2019} demonstrated that data augmentation techniques such as Cutout and Mixup are able to improve adversarial robustness.
From a theoretical perspective, Zhang et al.~\cite{Zhang:2021} proved that Mixup minimizes the adversarial loss on adversarial images and operates as a form of regularization on clean images. Thus, it can theoretically improve both robustness and generalization.

%
The seminal work of Geirhos et al.~\cite{Geirhos:2019} observed that convolutional networks trained on ImageNet exhibit a texture bias that decreases their representative performance.
They demonstrated that augmenting training samples with style transfer (replacing the texture from an image with another) alleviates texture bias and increases adversarial robustness.
Improving upon this idea, Shi et al.~\cite{Shi:2020} mitigated the texture bias of convolutional networks by removing local textures from the layer's output (i.e., feature maps). For this purpose, the authors adopted a dropout-based approach in which flat and high-frequency regions of a feature map have a higher probability to be dropped (zeroed out).
%
From extensive experiments, Mummadi et al.~\cite{Mummadi:2021} disproved that removing texture bias increases adversarial robustness. Instead, they confirmed that robustness comes from the data augmentation by the style transfer.
%
Li et al.~\cite{Yingwei_Li:2021:ICLR} showed that debiasing only texture or shape information degrades network performance on clean images. 
To improve generalization and robustness, they proposed a label assignment strategy that explores both texture and shape.

Closely related to data augmentation, increasing data training by combining clean and adversarial images, named adversarial training, is a successful defense mechanism~\cite{Madry:2018, Pang:2021}. It is worth mentioning that Cutout, MixUp and their variations naturally implement adversarial training on the occlusion attack, as they perturb regions of the image.

Despite the positive results, adversarial training requires careful regularization to avoid performance degradation on clean images~\cite{Chun:2019, Raghunathan:2020, Xie:2020:CVPR, Xie:2020, Bai:2021}.
Recent investigations revealed that adversarial training leads to inaccurate estimation of the batch normalization layers~\cite{Xie:2020, Xie:2020:CVPR, Wang:2020}, which increases the gap between generation and robustness.
%
As demonstrated by Xie et al.~\cite{Xie:2020:CVPR}, this problem takes place due to mismatching distribution between clean and adversarial images.
Furthermore, Xie and Yuille~\cite{Xie:2020} demonstrated that adversarial training demands very deep networks to obtain better robustness. 
Since deeper models are computationally inefficient, obtaining adversarial robustness by these mechanisms might limit applicability on resource-constrained devices.
Fortunately, we show an opposite phenomenon to Xie and Yuille~\cite{Xie:2020}, in which shallower networks (produced through removing layers) can provide gains in robustness.  

In contrast to adversarial training and other data augmentation techniques, we show that it is possible to improve robustness considering only clean images. 
Thereby, we can avoid assumptions of the attack and careful regularization.
%
Previously, Xie et al.~\cite{Xie:2018} reached this advantage by adding random scale and padding to an image before forwarding it through the network.
%
Unfortunately, their strategy incurs a higher computational cost since expanding scale is one factor that negatively impacts the computational performance of a network~\cite{Tan:2019:ICML, Kai:2020}.

\noindent
\textbf{Pruning Structures.} Pruning consists of locating and removing the least important structures (filters or layers) from convolutional networks preserving their predictive ability as much as possible.
To satisfy such conditions, previous works have demonstrated promising results using data-driven and data-agnostic criteria~\cite{Luo:2019, Tan:2020, Luo:2020, Jordao:2020, Lin:2020, Huang:2018:ECCV, Chin:2020, Guo:2020}.
%
While the first measures the importance of a structure according to its relationship with the data, e.g., using its features maps or loss, the latter estimates importance directly on the network parameters.
%
%

Most research on pruning aims at removing small structures such as filters or even weights. 
Veit et al.~\cite{Veit:2016} observed that residual architectures exhibit negligible accuracy drop when removing some of their layers\footnote{For simplicity, we use the term layers, but the work by Veit et al.~\cite{Veit:2016} confirmed such behavior when removing residual blocks -- sets of layers where the input of the first layer is added to the output of the last layer~\cite{He:2016}.}.
%
Building upon this, recent strategies are concentrated on removing entire layers~\cite{Wang:2018, Wu:2018, Zhang:2019, Veit:2020, Fan:2020}. Importantly, this family of pruning considers residual-based architectures only, as the evidence by Veit et al.~\cite{Veit:2016} is invalid to plain networks, e.g., VGG.

Regardless of the criterion for assigning importance or structure removed, pruning has been confirmed as a form of regularization~\cite{Huang:2016, Li:2019:CVPR, Fan:2020, Bartoldson:2020},  which reduces overfitting and improves generalization. In this work, we leverage this behavior through the lens of adversarial images, demonstrating its effect on robustness.
%
%
%
Concurrently to our work, Ye et al.~\cite{Ye:2019} demonstrated that adversarial training jointly with pruning leads to adversarial robustness. Hence, their work inherits all the drawbacks of adversarial training, for example, careful regularization and retraining from scratch when the attack is changed. On the other hand, our analysis prevents adversarial training, as it takes into account only clean images. Moreover, we conduct a more comprehensive evaluation since we investigate several pruning settings such as different schemes to parameters adjustment, structures removed, and pruning criteria. 
\section{Experiments}\label{sec:experiments}
\noindent
\textbf{Experimental Setup.}
Unless stated otherwise, we prune the networks using one iteration of pruning ($k=1$ in Algorithm~\ref{alg::pruning_structures}). We shall see that multiple iterations of pruning marginally influence the relationship between robustness and generalization.
To identify unimportant structures, we mainly focus on the Partial Least Squares (PLS) criterion~\cite{Jordao:2020}. However, we also demonstrate the influence of pruning on robustness when the pruning algorithm considers other criteria.
%
After pruning is complete, we fine-tune the resulting architecture for $200$ and $12$ epochs on CIFAR-10 and ImageNet, respectively. In this step, we use random crop and horizontal random flip as data augmentation~\cite{He:2016}. We also employ this data augmentation during the training phase (i.e., before running the pruning).
Such a setting plays an important role since it ensures we see adversarial images during the evaluation only.

To generate adversarial images, we employ the Fast Gradient Sign Method (FGSM) and model-agnostic attacks.
\begin{table*}[!!htb]
	\centering
	\small
	\renewcommand{\arraystretch}{1.2}
	\caption{Difference between accuracy of the unpruned (defenseless) network and its pruned counterpart. The symbols (+) and (-) indicate improvement and degradation, respectively, regarding the unpruned network. Last column shows the average improvement. Overall, the process of removing structures from convolutional networks provides robustness without degrading generalization, which reveals that pruning is capable of satisfying the dilemma between these metrics.}
	\label{tab:main_results}
	\begin{tabular}{ccccccc}
		\hline
		Architecture & Structure & Semantic & Occlusion & FGSM     & Clean    & Average  \\ \hline
		\multirow{3}{*}{ResNet56}    &  Filters & (+) \textbf{1.58} & (+) 2.76 & (+) \textbf{3.68} & (+) 0.60 & (+) \textbf{2.15} \\
		&  Layers  & (+) 1.05 & (+) 1.06  & (+) 3.20 & (+) 0.84 & (+) 1.53 \\
		&  Both    & (-) 4.13 & (+) \textbf{4.62}  & (+) 0.36 & (-) 0.60 & (+) 0.06 \\ \hline
		\multirow{3}{*}{MobileNetV2} &  Filters & (-) 0.60 & (+) \textbf{3.35} & (+) 0.64 & (+) \textbf{0.37} & (+) 0.94 \\
		&  Layers  & (-) 0.49 & (+) 2.12  & (+) \textbf{1.44} & (+) 0.15 & (+) 0.80 \\
		&   Both   & (+) \textbf{0.07} & (+) 2.56 &  (+) 1.05 & (+) 0.17&  (+) \textbf{0.96}        \\ \hline
		VGG16 & Filters & (-) 1.0  & (+) 4.86  & (-) 2.21 & (+) 0.89 & (+) 0.63 \\ \hline
	\end{tabular}
\end{table*}
%
In the latter, we use the semantic-preserving transformations (semantic for short) proposed by Hendrycks et al.~\cite{Hendrycks:2019:ICLR} and the occlusion attack for which we fill the image center with a square matrix of zeros, as suggested by previous works~\cite{Chun:2019, Yun:2019}. Unless otherwise specified, we consider the highest level of severity to the semantic-preserving ($\delta=4$) and occlusion ($\delta=16$), attacks and $\delta=8/255$ to FGSM.
We explore these attacks on plain (VGG16), lightweight (MobileNetV2) and residual architectures (ResNet56 and ResNet50). We report the performance in robustness and generalization using the difference, in percentage points (p.p), between the unpruned and pruned network. Thus, positive values, (+), indicate gains while negative values, (-), indicate degradation. All experiments were run on a machine with 16GB RAM and a single NVIDIA GTX 1080.
%
%
%

\noindent
\textbf{Robustness and Generalization from Pruning.} 
Our first experiment analyzes the effect of pruning on the trade-off between generalization (accuracy on clean images) and adversarial robustness (accuracy on adversarial images). Table~\ref{tab:main_results} summarizes the results on CIFAR-10. 
From this table, we observe that the networks become more robust to adversarial images after removing some of their structures (filters, layers or both).
For example, after pruning the filters of ResNet56, its robustness improved up to $3.68$ p.p. 
In particular, considering all the attacks we evaluate, pruning improved at least $1.58$ p.p the robustness of ResNet56.
The same trend occurred when pruning its layers, but with a smaller improvement in adversarial robustness. More concretely, pruning this structure, the gains in robustness ranged from $1.05$ p.p and $3.20$ p.p. It is worth noting that pruning provided these gains without degrading network generalization. On the other hand, a different behavior takes place when we prune both filters and layers.
On semantic-preserving, pruning both structures decreased robustness by $4.13$ p.p. We believe that the regulation severity caused by pruning both structures of ResNet56 impaired its performance.
In summary, despite the positive results on other attacks, the average improvement in pruning both structures is no better than removing filters and layers separated.
%

We observe that pruning MobileNetV2 and VGG16 is also an effective way of increasing their robustness.
%
%
%
By comparing the gains of pruning these architectures to ResNet56, we note that the average improvement decreased as a function of network depth. For example, the highest average improvement achieved by pruning ResNet56 (56-layer deep) was $2.15$ p.p, while on MobileNetV2 (51-layer depth) and VGG16 (16-layer deep) this improvement was $0.96$ and $0.63$ p.p.
%
As suggested by previous works~\cite{Wang:2018, He:2019, He:2020}, pruning brings better benefits to over-parameterized (i.e., deeper) networks. Our results corroborate this finding in the context of adversarial robustness.

Overall, the empirical evidence above shows that pruning improves robustness without sacrificing generalization, which is a desirable condition to any defense mechanism. Importantly, the pruning procedure takes into account clean images only. In contrast to previous research~\cite{Xie:2020, Xie:2020:CVPR}, which states that effective defense mechanisms require deeper networks, our results by pruning layers suggest an opposite phenomenon -- shallow networks can be as robust as their (unpruned) deeper counterpart.

\noindent
\textbf{Relationship between Generalization and Robustness.} 
It is well-known that pruned networks often achieve better generalization than their unpruned counterpart. Table~\ref{tab:main_results} reinforces this claim, in which removing $10\%$ of the filters of ResNet56 provides a generalization gain of $0.60$ p.p. Furthermore, a pruned network yielded by the $(k+1)$-th iteration of pruning might achieve higher generalization than its preceding (i.e., $k$-th) version.
According to Hendrycks et al.~\cite{Hendrycks:2019:ICLR} and Pang et al.~\cite{Pang:2021}, convolutional networks with higher generalization incur better robustness. Therefore, it is intuitive to assume that there exists a relationship between generalization and robustness.
In this experiment, we show that such a relationship exhibits a weak correlation from the perspective of pruning.

\begin{figure}[!b]
	\centering
	\includegraphics[width=\columnwidth]{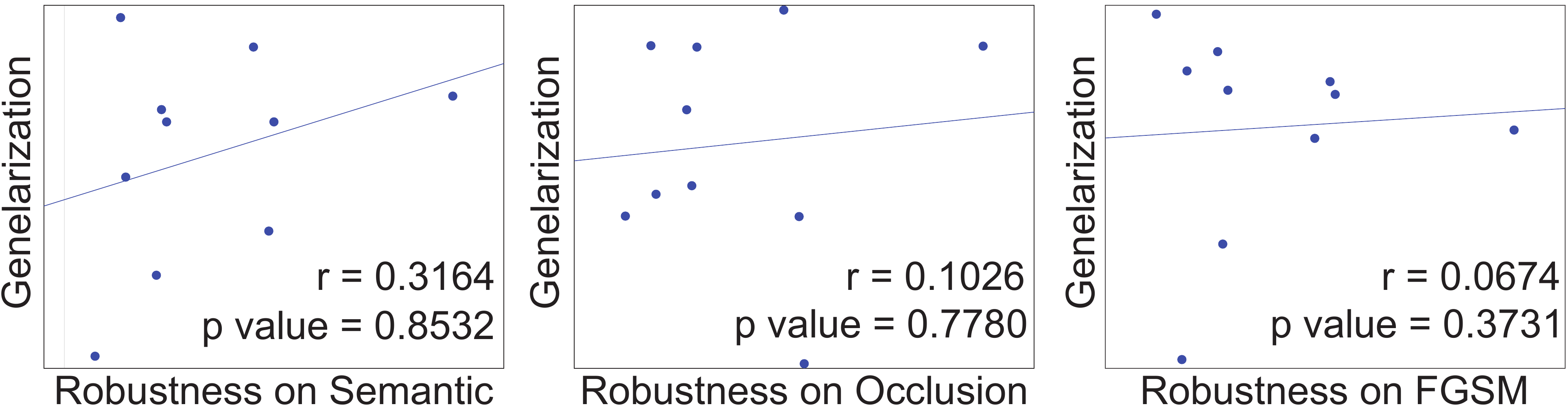}
	\caption{Relationship between generalization and robustness of pruned networks. Each point within the plot represents a pruned network yielded throughout the iterations of pruning; thus, $k$ points mean $k$ iterations of pruning. On different attacks, generalization and robustness are uncorrelated, implying that pruned networks with higher generalization do not incur better robustness.}
	\label{fig:correlation}
\end{figure}

To confirm the aforementioned statement, we ran several iterations of pruning and plotted the relationship between generation and robustness of each pruned network in Figure~\ref{fig:correlation}. It is evident from this figure that there is a weak relationship between generalization and robustness. To further validate this, we compute the Pearson's correlation coefficient ($r$) and notice that the coefficient is below $0.5$ on different adversarial attacks. In particular, the highest correlation coefficient was $0.31$, which indicates a weak correlation. The results from Table~\ref{tab:main_results} reinforce these observations, however, from the viewpoint of the structure removed.
%
 

These results show that pruned networks that obtain high generalization are not always the best candidates for achieving better robustness. Fortunately, we observe that one-shot pruning ($k=1$ in Algorithm~\ref{alg::pruning_structures}) provides a network capable of achieving competitive adversarial robustness. Since $k$ iterations of pruning mean $k$ stages of fine-tuning (see Algorithm~\ref{alg::pruning_structures}), one practical benefit of one-shot pruning is that we can avoid multiple epochs of fine-tuning.
\begin{table*}[!ht]
	\centering
	\small
	\renewcommand{\arraystretch}{1.2}
	\caption{Gains in robustness and generalization provided by removing structures of ResNet56 based on different pruning criteria. Differently from the standard assessment metric -- accuracy on clean images -- evaluating the pruning criteria on adversarial images reveals a shift in their effectiveness. In other words, top-performance criteria for clean images might perform poorly on adversarial images.}
	\label{tab:pruning_criteria}
	\begin{tabular}{cccccc}
		\hline
		Pruning Criterion & Semantic& Occlusion & FGSM     & Clean    & Average     \\ \hline
		$\ell_1$-norm               & (+) \textbf{1.64} & (-) 0.80  & (+) 3.75 & (+) 0.38 & (+) 1.24 \\
		ExpectedABS~\cite{Tan:2020} & (+) 0.96 & (-) 0.09  & (+) \textbf{4.29} & (+) 0.51 & (+) 1.41 \\
		HRank~\cite{Lin:2020}       & (+) 0.93 & (+) \textbf{2.92}  & (+) 3.18 & (+) 0.39 & (+) 1.85 \\
		KlDivergence~\cite{Luo:2020}& (+) 0.82 & (+) 0.73  & (+) 3.00 & (+) 0.34 & (+) 1.22 \\
		PLS~\cite{Jordao:2020}      & (+) 1.58 & (+) 2.76  & (+) 3.68 & (+) \textbf{0.60} & (+) \textbf{2.15} \\ \hline
	\end{tabular}
\end{table*}
%
%

\noindent
\textbf{Influence of the Pruning Criterion.} One key ingredient during the pruning process is the criterion for assigning structure importance. 
According to previous works~\cite{He:2020, Jordao:2020, Tan:2020}, the pruning criterion plays a role in preserving accuracy after pruning, as it decides which structures are irrelevant/redundant to the network. Depending on the strategy for measuring importance, the pruning criterion can also become a bottleneck in the computational cost of the pruning process~\cite{Luo:2020}. 
In this experiment, we investigate the behavior of the pruning criterion on adversarial robustness. 

Table~\ref{tab:pruning_criteria} shows the robustness of (pruned) ResNet56 when the pruning process employs different criteria for deciding which filters to remove. On the semantic-preserving and FGSM attacks, most pruning criteria provided similar gains in robustness. However, on the occlusion attack, the $\ell_1$-norm and the criterion by Tan and Motani~\cite{Tan:2020} performed poorly.
%
We shall see later that other pruning settings also affect the gains in the occlusion attack.

According to the average improvement (last column in Table~\ref{tab:pruning_criteria}), which includes gains in clean images, the best criterion was PLS.
%
In addition to these achievements, PLS exhibits other benefits, e.g., low memory consumption and execution time, which are attractive properties for resource-constrained scenarios. Such advantages are the reason we focus on PLS throughout our study. 
%
We highlight that all the pruning criteria evaluated are capable of improving robustness, except $\ell_1$-norm and ExpectedABS on the occlusion attack.
Thus, our main finding -- removing structures from convolutional networks improves their robustness -- is unbiased by a specific combination of pruning parameters. 

Interestingly, the results in Table~\ref{tab:pruning_criteria} suggest exploring robustness as an additional metric for evaluating the pruning criteria. For example, considering clean images only, the difference between the best and worst criteria is within one p.p, which is assumed a negligible difference~\cite{Blalock:2020}. On the other hand, this difference increases when we consider the occlusion and FGSM attacks. Additionally, criteria performing well on some attacks have their results notably decreased on others (e.g., $\ell_1$-norm and KlDivergence~\cite{Luo:2020}), thus demonstrating an unexpected instability. Thereby, we could employ robustness as an additional metric for evaluating the effectiveness of the pruning criteria, complementing previous efforts on improving the pruning evaluation~\cite{Blalock:2020}.

\noindent
\textbf{Effect of Pruning a Single Element.} In all the experiments so far, we analyze the effect of pruning on adversarial images when we remove a percentage (i.e., a small subset) of filters or layers followed by fine-tuning. In this experiment, we demonstrate that even removing a single element (i.e., only one filter) of a network influences their robustness.

To confirm the claim above, we perform the following process. First, we define $E$ as a set containing all possible\footnote{We highlight that some structures must remain unchanged (unpruned) due to technical details of the pruning process such as mismatching between feature maps.} filters or layers we could eliminate. Afterwards, we remove each element $e \in E$ individually and measure the generalization of the resulting network $\mathcal{F'}$. It is important to emphasize that $\mathcal{F'}$ corresponds to $\mathcal{F}$ without one element $e$.
Then, we add the elements whose removal provides a loss in generalization of $\{0, 1, 2, 3, 4, 5\}$ percentage points into a corresponding set $L_{i, i\in {\{0, 1, 2, 3, 4, 5}\}}$. Additionally, these elements are added into some $L_i$ only if they provide a gain in robustness.
For example, we insert the elements $e \in E$ that preserve the generalization (i.e., zero loss) into $L_0$ and the elements that drop generalization in up to one p.p into $L_1$. Roughly speaking, these steps allow us to evaluate our problem statement under different values of generalization and when the pruning process removes only one element.
\begin{algorithm}[!b]
	\small
	\caption{Pruning single Elements}
	\label{alg:pruning_single_element}
	\begin{algorithmic}
		\STATE {\bfseries  Input:} Unpruned Network $\mathcal{F}$
		\STATE {\bfseries  Input:} Elements to Prune $S$ (filters or layers)
		\FOR{$i \in \{0, 1, 2, 3, 4, 5\}$}
		\STATE $L_{i} \leftarrow \emptyset $
		\STATE $\sigma \leftarrow Acc_{clean}(\mathcal{F})-i$ $\triangleright$ Degradation in Generalization
		\FOR{$e \in E$ }
		\STATE $\mathcal{F'} \leftarrow \mathcal{F} \setminus e $ $\triangleright$ Removes $e$ from $\mathcal{F}$ 
		\IF{$Acc_{adv}(\mathcal{F'})$~$>Acc_{adv}(\mathcal{F})$~$\wedge$~$Acc_{clean}(\mathcal{F'})$~$>\sigma$}
		\STATE $L_i \leftarrow L_i  \cup \{e\} $
		\ENDIF
		\ENDFOR
		\ENDFOR
	\end{algorithmic}
\end{algorithm}  
%
%
Algorithm~\ref{alg:pruning_single_element} summarizes these steps. Note that we prevent any adjustment in the network parameters after removing an element. Such a setting relies on the fact that removing single filters or layers
marginally affects generalization~\cite{Veit:2016, Li:2019:CVPR, Luo:2020} and enables Algorithm~\ref{alg:pruning_single_element} to be computationally efficient.
\begin{figure}[!t]
	\centering
	\includegraphics[width=\columnwidth]{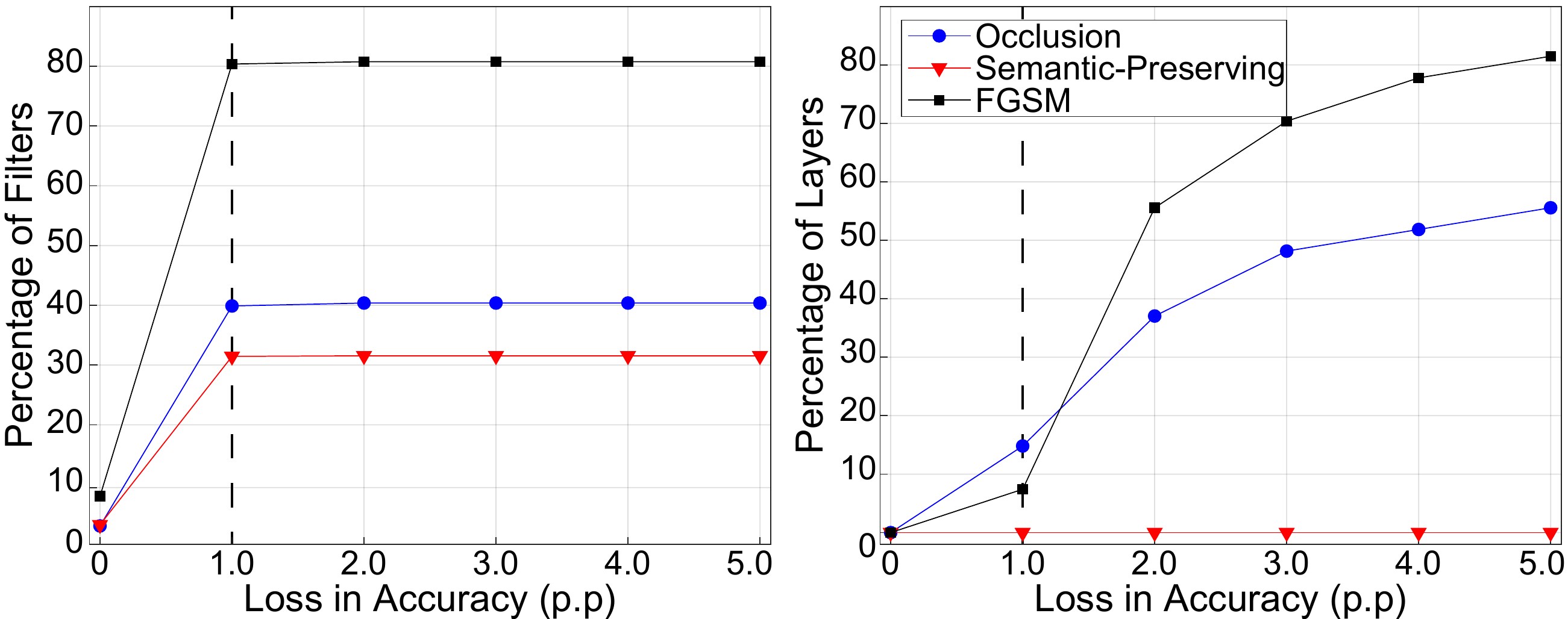}
	\caption{Percentage of structures ($y$-axis) that, when removed individually, increase robustness given a loss in generalization ($x$-axis). Take the FGSM attack as an example, with a negligible drop in generalization of one percentage point (dashed line) $80\%$ of the filters, if removed individually, are capable of achieving better adversarial robustness than the unpruned network. Importantly, these results consider removing a structure without fine-tuning.}
	\label{fig:IndividualElements}
\end{figure}

Figure~\ref{fig:IndividualElements} shows the relationship between the degradation in generalization and the percentage of elements ($\frac{|L_i|}{|S|}$) that, when removed individually, achieve better adversarial robustness than the unpruned network. 
%
Taking into account filters as elements in Algorithm~\ref{alg:pruning_single_element} and zero drop in generalization ($L_0$), we observe that less than $10\%$ of the filters yield a pruned network with gains in robustness.
This percentage substantially increases when we assume a drop in generalization of up to one p.p, for which it increased from $3.07\%$ to $39.88\%$ and from $8.04\%$ to $80.36\%$ on the occlusion and FSGM attacks, respectively. The curves rapidly saturate on higher values of degradation, indicating that no more filters can yield a pruned network with gains in robustness, even increasing degradation in generalization.

We observe a different behavior when considering layers as elements in Algorithm~\ref{alg:pruning_single_element}, Figure~\ref{fig:IndividualElements} (right).
For example, when removed individually, there was no layer capable of preserving generalization while improving adversarial robustness (it says $L_0$ is empty). Even evaluating the drop in $1$ p.p, the percentage of layers that improve robustness is below $20\%$ on all attacks. 

The previous analyses demonstrate that removing single elements without adjusting the network parameters enables us to obtain a certain degree of robustness. We can increase the number of elements satisfying these conditions by controlling the degradation in generalization. 
%
%
%
%

\noindent
\textbf{Training from Scratch vs. Fine-tuning.}
A broad discussion in pruning is how to adjust the network parameters after removing its structures. This adjustment is responsible for recovering the network's predictive ability and plays an important role in the computational cost of the pruning process. 
In this line of research, previous works either training the pruned network from scratch~\cite{Liu:2019, You:2020} or fine-tuning it for some epochs~\cite{Li:2019:CVPR, Luo:2019}. Regarding the first scheme, there are two distinct ways of assigning the network parameters: (i) randomly initialize the parameters~\cite{Liu:2019} and (ii) inherit the random initialization of the unpruned network -- winning tickets~\cite{Frankle:2019}.
%
In this experiment, we analyze the impact of these schemes on adversarial robustness. To training from scratch with parameters randomly initialized, following the setup by Liu et al.~\cite{Liu:2019}, we train a pruned network for the same number of epochs as its unpruned version (Scratch-B) and also consider doubling the epochs (Scratch-E). To the winning-ticket (W-ticket for short) scheme, we inherit the initialization from the unpruned network at epoch $50$.
We emphasize that all these schemes take into account only clean images.

\begin{table}[!t]
	\renewcommand{\arraystretch}{1.2}
	\centering
	\small
	\caption{Adversarial robustness considering different schemes to adjust the parameters of the pruned network. Scratch-E, Scratch-B and winning-ticket (W-ticket) train the pruned network from scratch (see the text for details) while fine-tuning adjusts the current parameters of the pruned network for a few epochs. Except for W-ticket, all schemes provide positive average improvements, with fining-tuning achieving the best gains.}
	\label{tab:comparison_adjustment}
	\begin{tabular}{ccccc}
		\hline
		& Semantic & Occlusion & FGSM  & Avg. \\ \hline
		Scratch-E   & (+) \textbf{1.72}  &  (-) 0.60      &  (+) 1.64          &  (+) 0.92   \\
		Scratch-B   & (+ )1.25           &  (-) 0.71       &  (+) 3.64          & (+) 1.39   \\ 
		W-ticket  & (+) 1.13       &    (-) 5.41   &  (+) 1.40  & (-) 0.96   \\ 
		Fine-Tuning &  (+) 1.58        & (+) \textbf{2.76} & (+) \textbf{3.68} &  (+) \textbf{2.67}   \\\hline
	\end{tabular}
\end{table}
Table~\ref{tab:comparison_adjustment} shows the gains in the robustness of ResNet56 using different schemes to adjust its parameters after the pruning process. In general, fine-tuning provided better improvements than the training from scratch schemes (Scratch-B/E and W-ticket). Interestingly, none of the training from scratch schemes was capable of providing gains on the occlusion attack. 
%
Fine-tuning, on the other hand, provided an improvement of $2.76$ p.p.
Overall, fine-tuning provided an average improvement of $2.67$ p.p., outperforming the best training from scratch (Scratch-B) by $1.28$ p.p. 
%

While Table~\ref{tab:comparison_adjustment} shows gains in robustness, we observe the same trend on generalization, where fine-tuning outperformed Scratch-B/E and W-ticket by $0.28$, $0.69$ and $0.01$ p.p, respectively. 
%
These experiments reinforce previous evidence that fine-tuning often leads to better generalization than training from scratch~\cite{Ye:2020}. Our analysis complements this hint, but for the robustness yielded by pruning.

\noindent
\textbf{Adversarial Transferability.} In previous experiments, we evaluate the adversarial robustness against FGSM using its standard setting --  a network is attacked by adversarial images crafted from its parameters. 
Here, we relax this assumption and assume that the FGSM attack is unaware of the victim (target) network.
%
To evaluate this transferability of adversarial images, we craft adversarial images from the parameters of a source network and attack a target network.
It is worth mentioning that the semantic-preserving and occlusion attacks are model-agnostic (i.e., they craft adversarial images without access to the network details). Therefore, such attacks were not used in this experiment.
\begin{table*}[!htb]
	\centering
	\small
	\renewcommand{\arraystretch}{1.2}
	\caption{Adversarial robustness when a model-specific attack (FGSM) yields adversarial images with different architectures including their pruned versions. Source and target indicate the network (to be precise its parameters) used to craft and evaluate the adversarial images, in this order. Thus, the off-diagonal shows robustness when the attack is unaware of the victim architecture or the defense mechanism (e.g., pruning). Bold values indicate the best robustness. Except when MobileNetV2 is the source, pruning an architecture provides the best adversarial robustness. These results are even better when the source and target architectures are different.}
	\label{tab:fgsm_tranferability}
	\begin{tabular}{ccccc}
		\hline
		\diagbox[width=8em]{Source}{Target} & ResNet56 & ResNet56 + Pruning & MobileNetV2 & MobileNetV2+Pruning \\ 
		\midrule
		ResNet56 & 21.60 & 32.73 & 52.55 & \textbf{53.25} \\
		ResNet56 + Pruning & 33.91 & 25.28 & 53.67 & \textbf{53.68} \\
		MobileNetV2 & \textbf{70.28} & 70.25 & 24.21 & 27.17 \\
		MobileNetV2 + Pruning & 69.69 & \textbf{70.01} & 28.29 & 24.84 \\ \hline
	\end{tabular}
\end{table*}
\begin{table}[!b]
	%
	\centering
	\small
	\renewcommand{\arraystretch}{1.2}
	\caption{Comparison of pruning with state-of-the-art defense mechanisms. Compared with competing defense mechanisms, pruning obtained one of the best average improvements. Such results indicate that pruning is capable of satisfying the dilemma between robustness and generalization. The symbols $\dagger$ and $\ddagger$ indicate results taken from Li et al.~\cite{Li:2021} and Chun et al.~\cite{Chun:2019}.}
	\label{tab:sota_imagenet}
	\begin{tabular}{cccc}
		\hline
		Defense      & Robustness & Generalization & Average  \\ \hline
		Stylized~$\dagger$      & (-) 2.29   & (-) 16.20      & (-) 9.24 \\
		MixUp~$\ddagger$                  & (-) 4.77   & (+) 1.10       & (-) 1.83 \\
		Cutout~$\ddagger$                 & (+) 1.39   & (+) 0.75       & (+) 1.06 \\
		CutMix~$\ddagger$                 & (+) 1.71   & (+) 2.07       & (+) 1.89 \\
		Shape-Texture~$\dagger$ & (+) 7.50   & (+) 0.50       & (+) 4.00 \\ \hline
		Pruning Filters & (+) 1.14   & (+) 3.15       & (+) 2.14 \\
		Pruning Layers & (+) 1.20   & (+) 3.03       & (+) 2.11 \\ \hline
	\end{tabular}
\end{table}

Table~\ref{tab:fgsm_tranferability} shows the results of the FGSM attack on different target and source architectures. From these results, when the source and target architectures are the same (diagonal -- the attack knows the victim), the adversarial robustness is below $30\%$. However, when the attack lacks access to the target architecture (off-diagonal), the adversarial robustness exhibited substantial gains.
Interestingly, the adversarial robustness is even better when the source and target architectures are completely different, for example, the source is MobileNetV2 (including its pruned version) and the target is ResNet56.
In addition, except when MobileNetV2 is the source, the pruned versions of the target architectures achieved even better results. Altogether, these results reinforce that pruned networks become more robust to adversarial images. Such evidence persists even when these images come from different architectures.

\noindent
\textbf{Comparison with Competing Defense Mechanisms.}
In this experiment, we compare the effectiveness of pruning with state-of-the-art defense mechanisms against adversarial images. It is worth mentioning that our focus is on showing the potential of pruning as a novel defense mechanism but not on pushing the state of the art in adversarial robustness. Following previous works~\cite{Chun:2019, Li:2021}, we evaluate several defense mechanisms on the ImageNet-C dataset~\cite{Hendrycks:2019:ICLR}. 

Table~\ref{tab:sota_imagenet} shows the gains (or degradation) in robustness and generalization of ResNet50 when we employ different defense mechanisms.
%
According to Table~\ref{tab:sota_imagenet}, the data augmentation by style transfer (Stylized for short) and MixUp exhibited degradation in the average improvement. 
Thus, these defense mechanisms fail to satisfy the dilemma between generalization and robustness. 
%
%
On the other hand, pruning provides gains in both metrics. Specifically, when the pruning process considers removing filters, the average improvement is $11.38$ and $3.97$ p.p higher than MixUp and Stylized, respectively. These gains are slightly smaller when the pruning process removes layers.
Compared to the defense mechanisms that simultaneously improve robustness and generalization (e.g., Cutout and CutMix), pruning achieved one of the best average improvements.
In particular, only the sophisticated data augmentation by Li et al~\cite{Li:2021} achieved a better average improvement than pruning.

It is worth mentioning that the mechanisms in Table~\ref{tab:sota_imagenet} require training the network from scratch. Pruning, however, obtained competitive results with only some epochs (i.e., $12$ epochs) of fine-tuning.
This advantage becomes pruning an attractive defense mechanism, mainly to off-the-shelf networks and scenarios with a limited training budget. Finally, since our pruning process employed only clean images, it is orthogonal to defense mechanisms based on data augmentation.  Therefore, we could combine pruning with these mechanisms, yielding more robust and efficient networks.
\section{Conclusions}\label{sec:conclusions}
We investigate the behavior of pruning through the lens of adversarial robustness.
We empirically show that pruning filters and/or layers of convolutional networks increase their adversarial robustness. Such evidence demonstrates that pruned networks do not inherit the adversarial vulnerability of their over-parametrized (unpruned) counterpart. Furthermore, pruning preserves generalization; thus, it efficiently satisfies the dilemma between robustness and generalization.
%
We confirm these findings considering only clean images during the pruning process. 
One practical benefit of this setup is to avoid adversarial training and, hence, its drawbacks. Additionally, our analysis enables us to design an effective defense mechanism that ignores the settings and assumptions of the attack. Compared to state-of-the-art defense mechanisms, pruning obtained one of the best trade-offs between robustness and generalization.
%
%
%
%
\section*{Acknowledgments}
The authors would like to thank FAPESP (grant \#2017/12646-3) and CNPq (grant \#309330/2018-1).
\bibliographystyle{ieee_fullname}
\bibliography{refs}

\end{document}